\documentclass{article}
\usepackage{ijcai26}
\usepackage{times}
\usepackage{soul}
\usepackage{url}
\usepackage[hidelinks]{hyperref}
\usepackage[utf8]{inputenc}
\usepackage[small]{caption}
\usepackage{graphicx}
\usepackage{subcaption}
\usepackage{amsmath}
\usepackage{amsfonts}
\usepackage{algpseudocode}
\usepackage{algorithm}
\usepackage{amsthm}
\usepackage{booktabs}
\usepackage{algorithm}
\usepackage{algpseudocode}
\usepackage[toc,page]{appendix}
\usepackage[switch]{lineno}
\usepackage{microtype}
\usepackage{dsfont}
\usepackage{bbm}
\usepackage{amssymb}
\usepackage{multirow}
\usepackage{verbatim}
\usepackage{microtype}

\usepackage{xcolor}

\newtheorem{definition}{Definition}

\title{Explaining Temporal Graph Predictions With Shapley Values}

\author{
Lea-Marie Sussek
\and
Stefan Heindorf
\affiliations
Paderborn University\\
Paderborn, Germany
\emails
\{lea.marie.sussek, heindorf\}@uni-paderborn.de,
}

\begin{document}

\maketitle

\begin{abstract}Temporal Graph Neural Networks (TGNNs) have become increasingly popular in recent years due to their superior predictive performance by combining both spatial and temporal information. However, how these models utilize the information to make predictions is rather unexplored, leading to potentially faulty or biased models. This work introduces two novel model-agnostic explainers for local explanations of TGNNs based on Shapley and Owen values. The first method, an event-level (edge-level) Shapley explainer, applies the KernelSHAP algorithm to estimate contribution scores for individual temporal events, providing interpretable descriptions for model behavior. The second, a feature-level Shapley explainer, extends this framework by decomposing event-level Shapley values into Owen values, and thereby uncovers hierarchical dependencies of the event and its features. The explainers outperform SOTA explainers on different metrics and datasets. Additionally, the Feature Explainer reveals a faulty extraction of actual timestamps of a commonly used TGAT implementation, helping to further understand performance drops on very sparse explanations. 
\end{abstract}

\section{Introduction}

Graph‑structured data appear in several domains such as social, biological, and communication networks. In the static setting, Graph Neural Networks (GNNs) effectively utilize structural information from nodes and edges. However, many real‑world graphs evolve over time, requiring models that combine temporal and structural information~\cite{wang2015link,luo2023care,vinchoff2020traffic}. Temporal Graph Neural Networks (TGNNs) do exactly this.

\cite{kazemi2020representation} classified temporal graph representations into Discrete‑Time Dynamic Graphs (DTDGs) and Continuous‑Time Dynamic Graphs (CTDGs). The former model the graph through temporal snapshots, but the models' predictive performance highly depends on the chosen granularity. The latter use a sequence of events, where each event is represented as a (source, destination, timestamp) triple. Despite remarkable progress, the models' internal decision-making processes are hard to grasp, rendering them effectively ``black boxes''. However, interpretability is essential for high-stakes domains such as healthcare, finance, and cybersecurity. Existing explainers for static GNNs fail to capture temporal effects~\cite{ying219gnnexplainer,vu2020pgm}, and existing explainers for discrete‑time TGNNs often depend on additional trainable models, making them slow and hyperparameter‑sensitive~\cite{he2022explainer,chen2023tempme,xia2023explaining}.

A promising alternative is to leverage Shapley values~\cite{shapley1953value,molnar2022interpretable}, which quantify how each input contributes to a model's output. Shapley‑based explanations are additive, contrastive, and do not require training auxiliary models which makes them appealing for model‑agnostic interpretability~\cite{molnar2022interpretable}. While such methods have been applied to static GNNs~\cite{akkas2024gnnshap}, their use for TGNNs remains largely unexplored.

We contribute two lightweight, model‑agnostic explainers for CTDGs based on Shapley theory. The \emph{Event Explainer} assigns Shapley values to each event without requiring training an auxiliary model, enabling compatibility with diverse TGNN architectures (e.g., attention‑based or random‑walk models). The values quantify the average contribution of an event to the model's prediction. The contribution describes both direction (positive or negative impact) and magnitude of influence (high or low impact). The \emph{Feature Explainer} is an extension of the first one by decomposing an event’s Shapley value into the contributions of the event's features, timing, and its function as an information transporter of neighboring nodes. To do so, we use Owen values~\cite{owen1977values,owen1995game}, an extension of Shapley values accounting for hierarchical structures. This extension provides fine‑grained insight into which event components mainly impact the Shapley value of the event.

Both explainers are computationally efficient, needing only a KernelSHAP~\cite{lundberg2017unified} approximation mixed with a Monte‑Carlo sampling approach. This results in a single hyperparameter needed for the explainer: the sample size. Evaluations on public benchmarks (MOOC, Wikipedia, Reddit) show that our methods achieve comparable or higher performance than state‑of‑the‑art explainers such as TempME and TGNN‑Explainer~\cite{chen2023tempme,xia2023explaining}, while also uncovering unexpected behaviors in attention-based models.

\section{Related Work}

\paragraph{Temporal Graph Neural Networks (TGNNs).}
Whereas TGNNs extend traditional GNNs by incorporating temporal information, Temporal Graph Attention Networks (TGATs) \cite{xu2020inductive} introduce multi‑head attention combined with a time encoder that encodes relative temporal distances between events. Although the attention weights provide some dense, low-level interpretability, they do not provide human-friendly, high-level  explanations~\cite{xia2023explaining}. Temporal Graph Networks (TGN)~\cite{rossi2020temporal} utilize a memory unit per node that is updated after each event and represents the history of a node in a compressed format. However, this usage of recurrent units limits parallelization. As an alternative, Causal Anonymous Walk Networks (CAWN)~\cite{wang2021inductive} perform random walks over time and base their decision on temporal motifs discovered.

\paragraph{Shapley‑based GNN Explainers.}
Several static GNN explainers use Shapley values to attribute importance to graph components. SubgraphX~\cite{yuan2021explainability} combines Monte‑Carlo tree search with Shapley values to identify subgraphs, whereas GraphSVX~\cite{duval2021GraphSVX} applies KernelSHAP to evaluate the influence of nodes and node features. Both treat explanation as a game where players (nodes or node's features) contribute to the model output.

\paragraph{Explainers for Temporal GNNs.}
Adapting the Monte-Carlo tree search of SubgraphX, TGNN-Explainer~\cite{xia2023explaining} uses a trainable navigator. In contrast to SubgraphX, TGNN-Explainer uses a reward based on prediction fidelity instead of Shapley values. TempME~\cite{chen2023tempme} focuses on identifying temporal motifs which are small, recurrent event sequences. Further, they use a dedicated GNN encoder and an information‑bottleneck training objective balancing accuracy and explanation compactness. Current explainers, however, rely on auxiliary models and deliver importance scores. This makes them hyperparameter sensitive, dependent on TGNNs that are differentiable w.r.t. event attentions, and costly to set up. Further, the importance scores do not quantify whether the event positively or negatively impacts the model's prediction. Therefore, this paper proposes two Shapley‑based explainers for CTDGs: a model‑agnostic Event Explainer attributing Shapley values directly to events, and a Feature Explainer leveraging Owen values to decompose an event's Shapley value into feature‑level effects, providing more in-depth explanations than existing approaches.

\section{Background}

A \textbf{Continuous-Time Dynamic Graph (CTDG)}~\cite{kazemi2020representation} models a temporal graph as a sequence of timestamped edges, called events. Formally, a CTDG is defined as $\mathcal{G} = (\mathcal V, \mathcal{E})$ with
\[
\mathcal{E} = \{(u, v, t, \mathbf x) \mid u,v \in \mathcal V,\; t \in \mathbb{R},\; \mathbf x \in \mathbb{R}^d\},
\]
where $\mathcal V$ denotes the set of nodes and $\mathcal{E}$ the set of events where an event $(u, v, t, \mathbf x) \in \mathcal{E}$ represents the edge from node $u$ to node $v$ at time $t$ and is accompanied by the event feature vector $\mathbf x \in \mathbb{R}^d$. 

The state of the graph at a specific timestamp is obtained by accumulating all events that occurred before that timestamp.

Temporal Graph Neural Networks (TGNNs) compute node embeddings $\mathbf h_{v,t}$ based on the \textit{computational subgraph} $G[v,t]=(V,E)$, where $V$ is restricted to the $k$-hop neighborhood of $v$ and $E$ is to the events with timestamps $t' < t$ within this neighborhood. Accordingly, $V\subseteq\mathcal V$ and $E\subseteq\mathcal E$. Events outside of $G[v,t]$ but belonging to the complete graph $\mathcal G$ are not considered during predictions.

The \textbf{Shapley value}~\cite{shapley1953value} provides a fair distribution of the total outcome of a cooperative game among its $N$ players. Let $\mathcal{N} = \{1,2,\dots,N\}$ be the set of players and let $val : \mathcal{P}(\mathcal{N}) \rightarrow \mathbb{R}$ be a value function assigning an outcome $val$ to each coalition of players, where $val(\emptyset)=0$. Then the marginal contribution of a player $i \in \mathcal{N}$ to a coalition $S \subseteq \mathcal{N}\setminus\{i\}$ is $val(S \cup \{i\}) - val(S)$. The Shapley value of player $i$ is defined as
\begin{align}
\label{eq:shapley}
\Phi_i = \sum_{S \subseteq \mathcal{N} \setminus \{i\}}
        &\frac{|S|!\,(|\mathcal{N}| - |S| - 1)!}{|\mathcal{N}|!}  \\
        &\cdot \big(val(S \cup \{i\}) - val(S)\big).\notag
\end{align}
This expression corresponds to the expected contribution of player $i$ over all possible joining orders of the players. It is the only distribution that satisfies the four properties of efficiency, symmetry, linearity, and null player~\cite{shapley1953value}. In particular, the efficiency property states that the Shapley values $\Phi_1, \dots, \Phi_N$ of individual players sum up to the value of the grand coalition of all players: $\sum_i \Phi_i = val(\mathcal{N})$.

In machine learning settings, the players correspond to input features and the value function measures the model's prediction relative to the average prediction.

When players are structured into groups that should be treated as units, the \textbf{Owen value}~\cite{owen1977values} generalizes the Shapley value. Let $\mathcal{T} = \{T_1, T_2, \ldots, T_k\}$ be a partition of $\mathcal{N}$ into disjoint groups, where $\bigcup_{j=1}^k T_j = \mathcal{N}$ and $T_i \cap T_j = \emptyset$ for $i \neq j$. For a player $i \in T_j$, the Owen value is defined as
\begin{align}
\label{eq:owen}
\Omega_i[\mathcal{T}] =
\sum_{H \subseteq \mathcal{T} \setminus \{T_j\}}
\sum_{S \subseteq T_j \setminus \{i\}}
w_H\cdot w_S \cdot
cont(i, val, H, S),
\end{align}
where $w_H=\frac{|H|!(k - |H| - 1)!}{k!}$ and $w_S=
\frac{|S|!(|T_j| - |S| - 1)!}{|T_j|!}$ are the weightings for the permutation of groups and the permutation within $i$'s group. $cont(i, val, H, S) = val(\cup_{H} \cup S \cup \{i\}) - val(\cup_{H} \cup S)$ accounts for the marginal contribution of $i$ with $\cup_{H}=\cup_{T\in H}T$.

An equivalent computation can be derived using the \emph{two-step} Owen values~\cite{owen1995game}. The Owen value of $i$ can be seen as the contribution of $i$ to the Shapley value of its group. Therefore, the Owen value is the Shapley value of $i$ in a game where the players are the members in $i$'s group, and the value function is the Shapley value of $i$'s group if only a subset of players is present. This hierarchical formulation preserves group structure and is particularly useful for model explanations where inputs are naturally grouped, such as features within events.

\section{Event-Level Explanations via Shapley Values}
\label{sec:event-expl}

Let $\mathcal{G}=(\mathcal{V},\mathcal{E})$ denote a temporal graph with nodes $\mathcal{V}$ and events $\mathcal{E}$. For a target node $v^*$ at timestamp $t$, the corresponding computational subgraph $G[v^*,t]=(V,E)$ contains the node and its time-restricted $L$-hop neighborhood such that all events occurred before $t$. Each event $e\in E\subseteq \mathcal{E}$ is represented as $e=(u,v,t,\mathbf x)$, where $u,v\in V\subseteq \mathcal{V}$ are source and destination nodes, $t\in\mathbb{R}$ is the event timestamp, and $\mathbf x\in\mathbb{R}^d$ denotes the event's feature vector. A trained TGNN is defined as $f: \mathcal{G}\rightarrow \mathbb{R}$, producing a prediction $f(G[v^*,t])$ for a computational subgraph $G[v^*,t]$.

To obtain event-level attributions, we formulate a cooperative game in which the \emph{players} are all events $E$ in the computational subgraph, and the \emph{value function} quantifies their contribution to the model output:
\begin{equation}
val(S) = f(m(S,G)) - f(m(\emptyset,G)), \quad S\subseteq E,
\label{eq:value-function}
\end{equation}
where $m(S,G)$ denotes a masked subgraph that keeps only the events in $S$. Since removing events may disconnect the graph \cite{duval2021GraphSVX}, masked events are replaced by their corresponding \emph{average event} $\bar e$.

\begin{definition}{Average Event.}
For an event $e=(u,v,t,\mathbf x)$ and the timestamp $t_v$ at which we observe $e$,
the average event is defined as
\begin{equation}
\bar e = (u,v,t_v-\bar t,\mathbf {\bar x}),
\end{equation}
where $\mathbf {\bar x}\in\mathbb{R}^d$ is the component-wise mean of all event feature vectors in $\mathcal{E}$, 
and $\bar t$ is the mean temporal difference between consecutive events.
\end{definition}

$t_v$ is usually the timestamp at which we make a prediction or the timestamp at which we reached node $v$. The temporal adjustment in the average event ensures consistency with TGNN encoders that rely on time deltas rather than absolute timestamps (e.g., TGAT~\cite{xu2020inductive}). The complete masking function is then
\begin{equation}
m(S,G) = (V, E'), \quad
E' = S \cup \{\,\bar e \mid e\in E,\ e\notin S\,\}.
\end{equation}
To investigate the event's impact as an event that only propagates information from a more distant neighborhood, we also set node features of the source node $u$ to zero if the event is removed. Consequently, this requires the TGNN to initialize node features with the zero vector. In contrast to the TGNN-Explainers~\cite{xia2023explaining} that explores connected subgraph structures, we explore disconnected subgraph structures. Therefore, using an average event instead of removing the event is crucial to avoid side-effects from disconnected parts.

\begin{figure*}[htbp]
    \centering
    
    % Subfigure 1
    \begin{subfigure}[b]{0.3\textwidth}
        \centering
        \includegraphics[width=\textwidth]{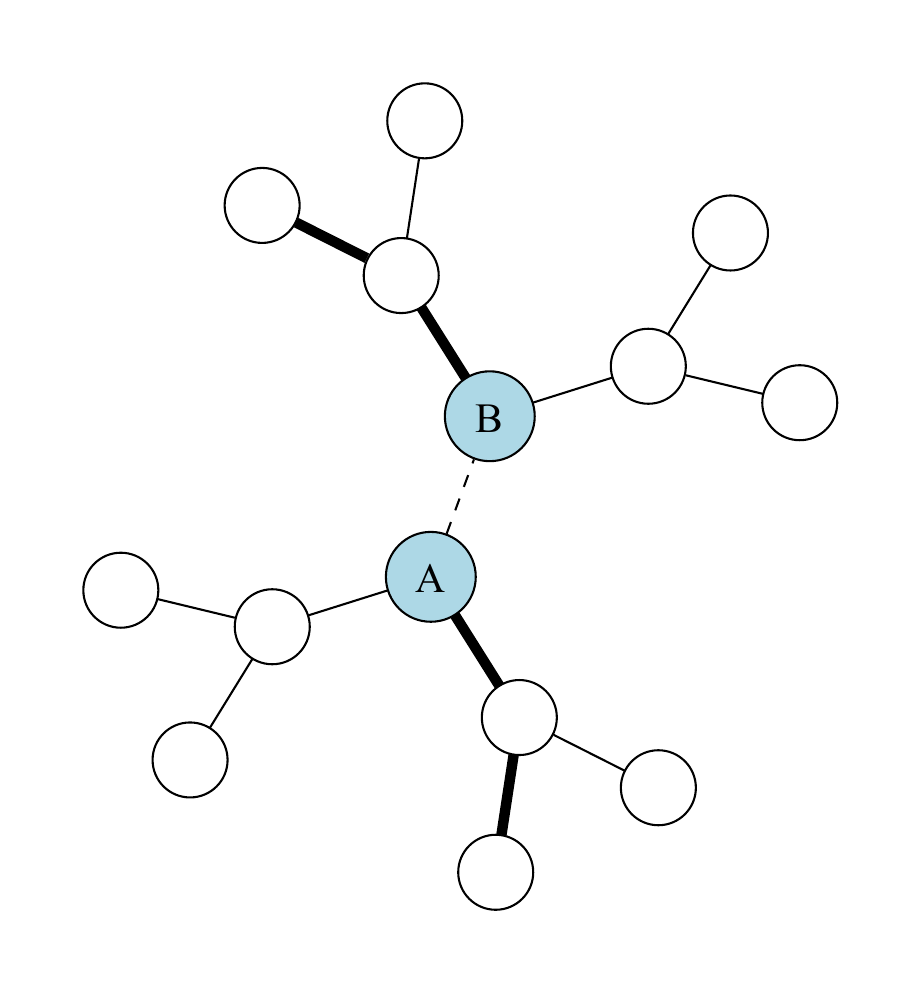}
        \caption{TGNN-Explainer.}
        \label{fig:sub1}
    \end{subfigure}
    \hfill
    % Subfigure 2
    \begin{subfigure}[b]{0.3\textwidth}
        \centering
        \includegraphics[width=\textwidth]{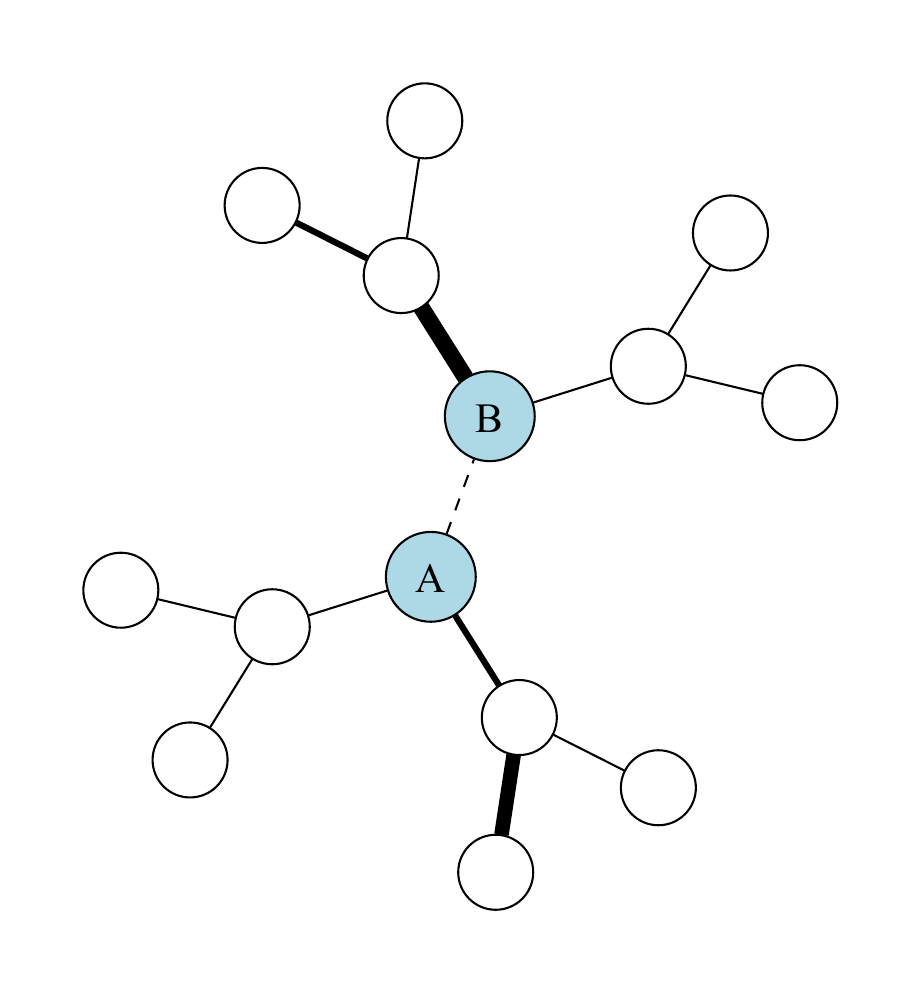}
        \caption{TempME.}
        \label{fig:sub2}
    \end{subfigure}
    \hfill
    % Subfigure 3
    \begin{subfigure}[b]{0.3\textwidth}
        \centering
        \includegraphics[width=\textwidth]{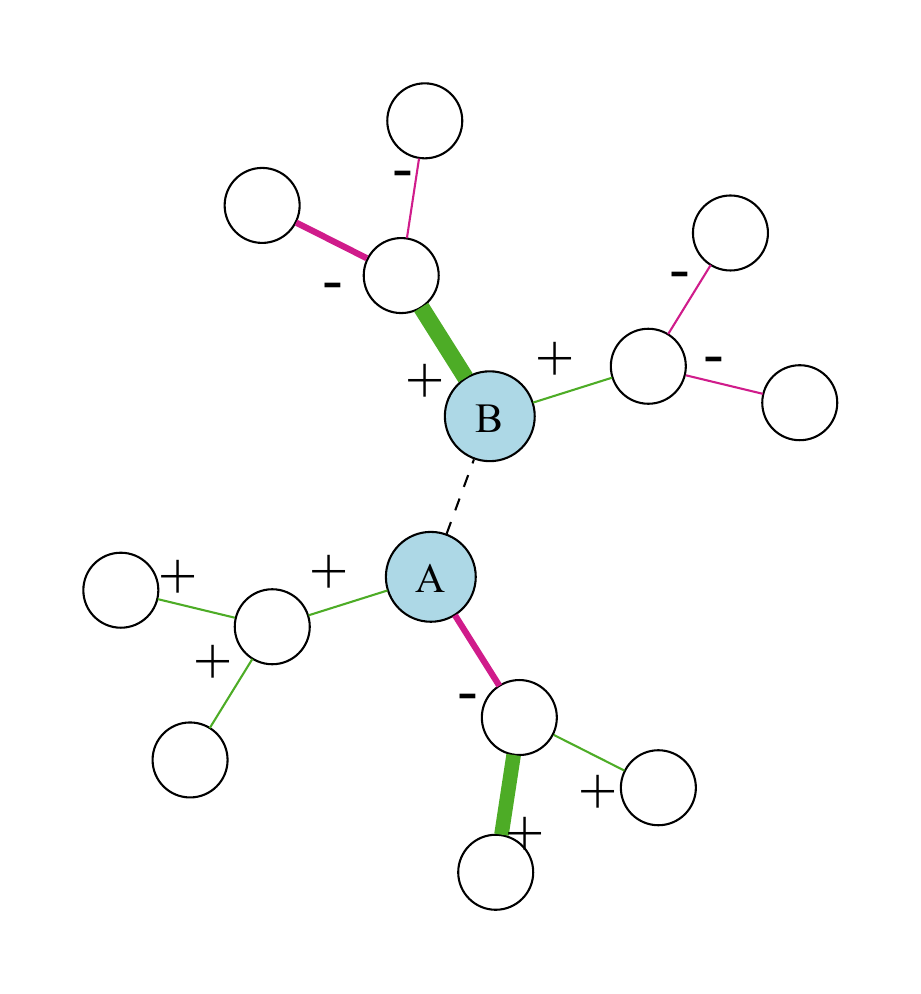}
        \caption{TGNNSHAP (ours). }
        \label{fig:sub3}
    \end{subfigure}
    
    \caption{Comparison of the explanations generated by the different explainers. The connection between $A$ and $B$ is predicted and the prediction needs to be explained. TGNN-Explainer returns a subset of important connected events. TempME assigns each event an importance score. The thickness of the event indicates its importance. The Event Explainer (ours) quantifies the average contribution to the prediction. Purple and green means positive or negative influence, respectively. The thicker the line the more dominant is the influence.}
    \label{fig:expl-comparison}
\end{figure*}

\paragraph{Approximation of Shapley Values.}
Exact computation requires $O(2^{|E|})$ evaluations, becoming intractable for even moderately sized subgraphs. Further, the number of events within the computational subgraph grows exponentially in the number of hops $L$. We therefore apply KernelSHAP approximation~\cite{lundberg2017unified}, which fits a weighted linear model $g(z)=\Phi_0+\sum_{i=1}^{|E|}\Phi_i z_i$ to approximate the Shapley values using a predefined number of samples. $z_i\in\{0,1\}^{|E|}$ is the coalition vector as defined in~\cite{lundberg2017unified}. To this end, each event receives a Shapley value quantifying its average contribution to the prediction (see Fig.~\ref{fig:expl-comparison}).

\section{Feature-Level Explanations via Owen Values}
\label{sec:featureexplainer}

While event-level explanations based on Shapley values provide useful insights into the importance of events, they are too coarse to analyze the importance of \emph{event features}. Hence, our \textit{Feature Explainer} adds a finer granularity by decomposing each event into its constituent features.

\paragraph{Definition of Game.}
 Let $E=\{e_1,\ldots, e_{|E|}\}$ denote the events in the computational subgraph $G[v^*,t]$ and $F'=\{1,\ldots,|F'|\}$ the event features. In addition, the event timestamp $t_v$ and the corresponding source node $u$ of the event $e$ are considered as two additional players per event, resulting in $|E|\cdot(|F'|+2)$ total players within the computational subgraph where $|F'|$ is the number of event features per event. For a more compact notation, we call the two additional players also features, resulting in $|F|=|F'+2|$ features. The source node is considered as player to investigate whether a node only serves as ``information bridge'' to propagate information from far-away nodes. The target node features are not relevant for our datasets as node features are always zero.

 The value $val:\mathcal{P}(E \times F) \rightarrow \mathbb{R}$ of a coalition of players is obtained akin to Eq.~\ref{eq:value-function} by making a prediction for node $v$ at time $t$. However, we now consider subsets of fine-granular players $S\subseteq E \times F$ where in the computation graph $G$, events $e=(u, v, t_v, x_1, \ldots, x_d)$ can be missing partially: if the time stamp $t_v$ is missing, it is replaced by $t_v - \bar{t}$; if a single feature value $x_i$ is missing, it is replaced by its (column-wise) mean $\bar{x}_i$ across all events $e \in E$; if the ``source node feature player'' is not present, the node features of the source node are set to zero in the computation graph. The obtained computation graph $G'$ is used to make a prediction $f(G'[v^*,t])$. This construction ensures that the resulting graph remains connected and the TGNN can make a prediction for instances with missing data.

\paragraph{Limitations of Shapley Values.}
Computing Shapley values $\Phi_f$ for each feature player suffers from two major issues: Computational complexity and inconsistency with the Shapley values calculated of by the Event Explainer. Having $|E|\cdot|F|$ many players, leads to the evaluation of $O(2^{|E|\cdot |F|})$ many coalitions, which quickly becomes infeasible for realistic graphs. The inconsistency roots from the fact that the efficiency property of Shapley values does not hold for arbitrary subsets of players. Thus, if we sum the Shapley values of features for one event, we cannot guarantee to receive the Shapley value of the event. Formally, $\Phi_e = \sum_{f\in F} \Phi_f$ does not hold in general. Therefore, we reformulate the feature-level explanation as a hierarchical game, as described in the following.

\subsection{Hierarchical Explanation via Owen Values}

The hierarchical structure of features within events enables the use of \textit{Owen values}~\cite{owen1977values} that---as we will show in the following---preserve additivity across hierarchical levels. Let the set of event groups be $\mathcal{T}=\{T_1,\ldots,T_{|E|}\}$, where each group $T_e$ contains the feature-level players of event $e$. Formally, for a player (feature) $f\in T_e$, the Owen value is defined as

\begin{align}
\label{eq:owen-event}
\Omega_f[\mathcal{T}] =
\sum_{H \subseteq \mathcal{T} \setminus \{T_e\}}
\sum_{S \subseteq T_e \setminus \{f\}}
w_H\cdot w_S \cdot
cont(f, val, H, S),
\end{align}
where $w_H=\frac{|H|!(|E| - |H| - 1)!}{|E|!}$, $w_S=
\frac{|S|!(|F| - |S| - 1)!}{|F|!}$ are weightings for the permutation of events and permutation within $f$'s event group. $cont(f, val, H, S) = val(\cup_{H} \cup S \cup \{f\}) - val(\cup_{H} \cup S)$ is defined as before and accounts for the marginal contribution of feature $f$. Since $|F|=|T_e|$, Eq.~\ref{eq:owen-event} is the application of Eq.~\ref{eq:owen} using the event-wise grouping $\mathcal{T}=\{T_1,\ldots,T_{|E|}\}$.

Intuitively, $\Omega_f[\mathcal{T}]$ measures the expected marginal gain to the Shapley value of the event if the feature joins a coalition of other features belonging to the event.

The Owen values preserve a fundamental hierarchical property:
\begin{equation}
\label{eq:hier-eff}
\sum_{f\in T_e}\Omega_f[\mathcal{T}]
 = \Phi_{T_e}[val_{\mathcal{T}}],
\end{equation}
where $\Phi_{T_e}[val_{\mathcal{T}}]$ is the Shapley value of event $e$ in the game defined by $val_{\mathcal{T}}$ that treats each event as a single player, i.e. the game played by the Event Explainer. This property follows from the alternative ``two-step'' formulation \cite{owen1995game} which formulates the Owen value of a player as Shapley value of this player contributing to the Shapley value of the player's group. As shown in \cite{owen1995game}, this two-step formulation is equivalent to the standard definition of the Owen value. Under this interpretation, the hierarchical efficiency property in Eq.~\eqref{eq:hier-eff} becomes immediate: since the Owen value represents a player’s contribution to its group’s Shapley value, summing the Owen values of all players in a group yields the Shapley value of that group. Note that the Shapley values computed for the contribution of the players and those computed for the groups correspond to different underlying games. In the TGNN explanation context, this guarantees that the sum of a given event’s feature-level attributions matches its event-level contribution, ensuring consistency across explanatory levels.

\subsection{Approximation of Owen Values}

Computing Eq.~\ref{eq:owen-event} is expensive as it requires $\Theta(2^{|E|+|F|})$ model calls, and the number of events $|E|$ and features $|F|$ are often large. To make the computation tractable, we apply the ``two-step'' formulation of Owen values~\cite{owen1995game} using Shapley value approximations. The goal is to calculate the Owen values of all features of event $e$ efficiently.

We leverage the intuition behind the ``two-step'' formulation that the Owen values can be seen as Shapley values of Shapley values. In other words, they quantify how much each feature contributes to the Shapley value of $e$. Therefore, the features play a game whose value function is the Shapley value of $e$ (as in Sec.~\ref{sec:event-expl}, but only a subset of $e$'s features are present). Thus, the Owen values now become Shapley values, and we can use KernelSHAP to synchronously calculate them using $l$ samples. However, the game's outcome is the Shapley value of $e$ where $e$'s features are altered. Since we are only interested in the Shapley value of $e$, we can use Monte Carlo sampling with $k$ samples to obtain the Shapley value in the second step.

To accelerate the computation, we utilize parallel precomputation of the value function for different coalitions. To do so, we sample $k$ random coalitions $K$ of events excluding the event $e$ to be explained. Next, we sample $l$ coalitions of features $L$. Then, we compute the predictions $\hat {\mathbf y}_{-e}\in \mathbb R^{k}$ using only the subgraphs of our $k$ sampled coalitions, while the excluded events are replaced by their average events. Note, that event $e$ is absent for all predictions in $\hat {\mathbf y}_{-e}$. After that, we create the predictions $\hat Y\in \mathbb R^{l\times k}$ of all possible combinations of $K$ and $L$. This means that $\hat Y_{i,j}$ is the prediction on the subgraph using $e$ together with the events $K_j$ and $e$ has only the features in $L_i$. All other feature values are set to their average. We utilize Monte-Carlo sampling by element-wise subtracting $\hat {\mathbf y}_{-e}$ from each row in $\hat Y$ and calculating a row-wise mean resulting in $\Phi_L\in \mathbb R^l$. The $i$-th element in $\Phi_L$ now contains the approximated Shapley value of $e$ if the feature coalition $L_i$ is present. That means $\Phi_L$ contains the coalition values for the different coalition samples in $L$. Therefore, using these coalition values, we can use KernelSHAP~\cite{lundberg2017unified} to compute a Shapley value for each feature value. These Shapley values are the desired Owen values.

Consequently, the overall process yields an efficient, hierarchical approximation of Owen values with a complexity of $O(k \cdot l)$ instead of an exponential complexity in $|E|$ and $|F|$.

\section{Evaluation}

After briefly introducing our evaluation setup, we evaluate our explainers both quantitatively and qualitatively.

\subsection{Evaluation Setup}
\label{sec:eval-setup}

The proposed Feature and Event Explainers are systematically compared with the baseline explainers TGNN-Explainer and TempME. For TGNN-Explainer, we adopt the public implementation from \cite{ghasemire2024reproducibility}\footnote{\url{https://github.com/cisaic/tgnnexplainer}}. Its original expansion mechanism, which uses an MLP guided by attention weights, is limited to attention-based architectures such as TGAT. To ensure general applicability, we replace this step with the random-event expansion strategy already used by~\cite{xia2023explaining}. TempME is evaluated using the official codebase\footnote{\url{https://github.com/Graph-and-Geometric-Learning/TempME}} as proposed by \cite{xu2020inductive}. Its attention-based continuous masks are thresholded to obtain sparse sets of events according to target sparsity levels, using the provided or default configurations.

\paragraph{Datasets.} Evaluation is performed on three real-world temporal link prediction benchmarks—MOOC, Wikipedia, and Reddit~\cite{poursafaei2022towards}—as well as one synthetic dataset. MOOC captures student–page interactions, while Wikipedia and Reddit are bipartite graphs with 172 LIWC-based features~\cite{pennebaker2001linguistic}. The synthetic benchmark, implemented to isolate temporal effects, comprises 50 timestamps with 1{,}000 generated subgraphs each, using only zero-valued node and event features. Temporal graph models are drawn from the DyGLib collection~\cite{yu2023towards}, ensuring comparable baseline predictive accuracy to prior studies.

\paragraph{Metrics.} Explanation quality is quantified through four complementary measures averaged over 200 generated explanations per sparsity threshold $s\in[0,1]$:  
Fidelity, $-\!|\hat{y}-\hat{y}_{\mathrm{sparse}}|$, assessing the preservation of model output with $\hat y=\sigma(f(G))$ being the original prediction and $\hat y_{\mathrm{sparse}}=\sigma(f(G_{\mathrm{sparse}}))$ being the prediction on the sparse computational subgraph;  
Deviation, $-\!|y-\hat{y}_{\mathrm{sparse}}|$, quantifying proximity to the ground truth $y$; 
Logit-based Fidelity~\cite{chen2023tempme,xia2023explaining}, evaluating consistency at the logit level using $\mathbbm 1_{y=1}(f(G_{\mathrm{sparse}})-f(G))+\mathbbm 1_{y=0}(f(G)-f(G_{\mathrm{sparse}}))$ where $f(G)$ and $f(G_{\mathrm{sparse}})$ are the logit values of the TGNN; and
Graph Explanation Faithfulness (GEF)~\cite{agarwal2023evaluating}, computed as $\exp^{-\mathrm{KL}(\hat{y}\,\|\,\hat{y}_{\mathrm{sparse}})}$, which relies on Kullback–Leibler divergence to capture distributional similarity. 

\paragraph{Framework.} All explainers produce event- or feature-level relevances that are unified into sparse subgraphs for comparison. TGNN-Explainer outputs discrete event subsets, TempME ranks events via attention scores, while the Event and Feature Explainers rely on absolute Shapley and Owen values, respectively. To limit computational cost, the Feature Explainer is restricted to the top three events identified by the Event Explainer. The experiments were conducted in an HPC cluster.

\paragraph{Reproducibility.} Our experiments were conducted on an HPC cluster utilizing the NVIDIA A100 40 GB GPU and 80G RAM. The TGNNs and the baselines are configured as reported by ~\cite{yu2023towards,xia2023explaining,chen2023tempme}. If for a dataset no configurations were available, we used the default values. The code underlying our research is publicly available.\footnote{\url{https://github.com/ds-jrg/TGNNSHAP}} Further, we use the sample size for KernelSHAP provided by SHAP\footnote{https://shap.readthedocs.io/en/latest/} and a sample size of 550 for the Monte Carlo sampling. The latter was selected in an empirical pilot study as a good trade-off between runtime and explanation variance.

\subsection{Quantitative Evaluation}

\begin{table*}[!ht]
\centering
\setlength{\tabcolsep}{5pt}
\begin{tabular}{@{}llrrrrrrrr@{}}
\toprule
 & \textbf{Explainer} & \multicolumn{4}{c }{\textbf{Events are removed}} & \multicolumn{4}{c}{\textbf{Events replaced by mean}} \\
 \cmidrule(lr){3-6} \cmidrule(l){7-10}
&  & Fidelity & Fid. (Logit) & Deviation & GEF & Fidelity & Fid. (Logit) & Deviation & GEF \\
\midrule

% -------- Generated --------
\parbox[t]{7mm}{\multirow{4}{*}{\rotatebox[origin=c]{90}{Generated}}} & TGNN Explainer & -15.8074 & -- & -15.8138 & -- & -21.9696 & -- & -21.9758 & --\\
& TempME & --17.3761 & -- & -17.3826 & -- & -23.0079 & -- & -23.0142 & --\\
& TGNNSHAP (Event) & \underline{-6.2456} & -- & \underline{-6.2515} & -- & \underline{-7.0813} & -- & \underline{-7.0865} & -- \\
& TGNNSHAP (Feature) & \textbf{-0.2733} & -- & \textbf{-0.2773} & -- & \textbf{-0.0707} & -- & \textbf{-0.0766} & -- \\
\midrule

% -------- MOOC --------
\parbox[t]{7mm}{\multirow{4}{*}{\rotatebox[origin=c]{90}{MOOC}}} & TGNN Explainer & -0.3169 & -1.2321 & -0.5581 & 0.4228 & -0.2380 & \textbf{0.7680} & \textbf{-0.1713} & 0.7431\\
& TempME & -0.0791 & \underline{-0.1604} & -0.3708 & 0.8704 & -0.1193 & -0.3701 & -0.4114 & 0.7949\\
& TGNNSHAP (Event) & \textbf{-0.0671} & \textbf{-0.1431} & \underline{-0.3682} & \textbf{0.9057} & \underline{-0.0969} & \underline{-0.2321} & -0.3884 & \underline{0.8335}\\
& TGNNSHAP (Feature) & \underline{-0.0672} & -0.2019 & \textbf{-0.3376} & \underline{0.8864} & \textbf{-0.0767} & -0.2883 & \underline{-0.3519} & \textbf{0.9063}\\
\midrule

% -------- MOOC (TGN) --------
\parbox[t]{7mm}{\multirow{4}{*}{\rotatebox[origin=c]{90}{\shortstack{MOOC\\(TGN)}}}} & TGNN Explainer & -0.2100 & -1.0066 & -0.4278 & 0.7302 & -0.2829 & -1.4083 & -0.5026 & 0.6201\\
& TempME & -0.1395 & -0.6019 & -0.3659 & 0.8208 & -0.2507 & -1.1832 & -0.4657 & 0.6648\\
& TGNNSHAP (Event) & \underline{-0.0742} & \underline{-0.1879} & \underline{-0.3102} & \underline{0.9208} & \underline{-0.1148} & \underline{-0.2496} & \underline{-0.3256} & \underline{0.8565}\\
& TGNNSHAP (Feature) & \textbf{-0.0490} & \textbf{-0.0255} & \textbf{-0.2863} & \textbf{0.9514} & \textbf{-0.0853} & \textbf{-0.0853} & \textbf{-0.3010} & \textbf{0.8981}\\
\midrule

% -------- Wikipedia --------
\parbox[t]{7mm}{\multirow{4}{*}{\rotatebox[origin=c]{90}{Wikipedia}}} & TGNN Explainer & -0.1299 & \textbf{0.7639} & \textbf{-0.0917} & 0.8899 & -0.0864 & \textbf{3.6979} & \textbf{-0.0246} & 0.9343\\
& TempME & -0.0563 & -0.4377 & \underline{-0.1214} & 0.9474 & -0.0409 & \underline{0.3842} & \underline{-0.0901} & 0.9612\\
& TGNNSHAP (Event) & \underline{-0.0484} & -0.4242 & -0.1235 & \underline{0.9510} & \textbf{-0.0274} & 0.3390 & -0.0935 & \textbf{0.9764}\\
& TGNNSHAP (Feature) & \textbf{-0.0452} & \underline{-0.4086} & -0.1225 & \textbf{0.9556} & \underline{-0.0278} & 0.2950 & -0.0923 & \underline{0.9754}\\
\midrule

% -------- Reddit --------
\parbox[t]{7mm}{\multirow{4}{*}{\rotatebox[origin=c]{90}{Reddit}}} & TGNN Explainer & -0.2270 & -2.1331 & -0.2508 & 0.8683 & -0.1709 & -1.7020 & \textbf{-0.1602} & 0.8922\\
& TempME & \underline{-0.0823} & -0.7485 & \textbf{-0.2123} & 0.9063 & -0.0616 & \textbf{-0.7353} & \underline{-0.1610} & 0.9528\\
& TGNNSHAP (Event) & -0.0836 & \underline{-0.7114} & -0.2140 & \underline{0.9095} & \underline{-0.0592} & -0.8638 & -0.1683 & \underline{0.9559}\\
& TGNNSHAP (Feature) & \textbf{-0.0815} & \textbf{-0.6599} & \underline{-0.2131} & \textbf{0.9140} & \textbf{-0.0574} & \underline{-0.8629} & -0.1680 & \textbf{0.9584}\\

\bottomrule
\end{tabular}
\caption{Comparison of AUCs for different explainers. Left block: events removed (zeroed out). Right block: events replaced by mean event or mean feature value. Bold is the best performing explainer and the second best is underlined. For all metrics, higher means better.}
\label{tab:AUCs-combined}
\end{table*}

\begin{figure}
    \centering
    \includegraphics[width=0.85\linewidth]{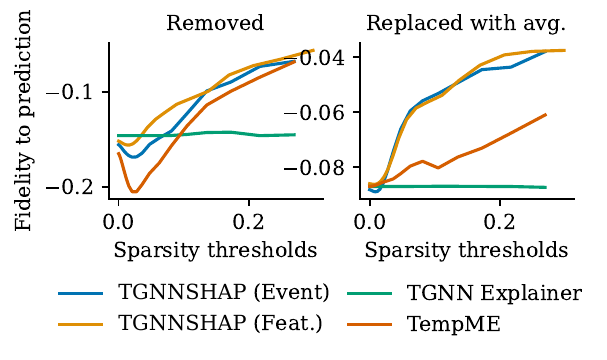}
    \caption{Lower part of the Sparsity-vs-Fidelity curve on the Wikipedia dataset using a TGAT model.}
    \label{fig:evalWikipedia}
\end{figure}

\begin{figure}
    \centering
    \includegraphics[width=0.85\linewidth]{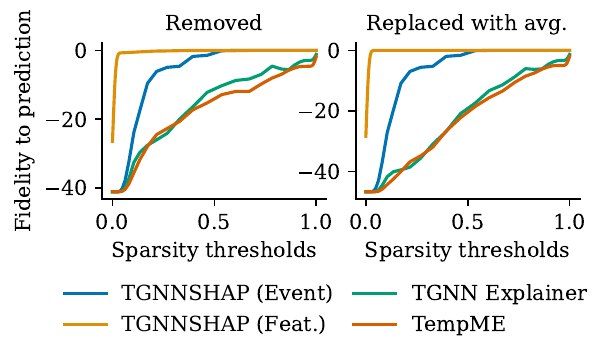}
    \caption{Complete Sparsity-vs-Fidelity curve on the generated dataset using a TGAT model.}
    \label{fig:evalGen}
\end{figure}

The comparison encompasses multiple datasets and all metrics defined earlier, with area-under-the-curve (AUC) scores computed from the Sparsity-vs-Metric curves using the full range of sparsities. The complete results are reported in Tab.~\ref{tab:AUCs-combined}, while representative Sparsity-vs-Metric curves are shown in Fig.~\ref{fig:evalWikipedia} and~\ref{fig:evalGen}. We conducted our experiments by performing two kinds of event removal: Excluding the event with the risk of creating disconnected graphs and replacing an event by its average event to maintain connectivity. However, using average events preserves structural information like certain shapes or the fact that two node communicated independent of the time stamp.

Across all datasets, the proposed explainers achieve higher AUC values for most metrics, demonstrating superior explanatory power relative to TGNN‑Explainer and TempME. The average rank of the Feature Explainer and the Event Explainer is 1.58 and 2.19, respectively. TempME and the TGNN-Explainer are on average on rank 2.92 and 3.31, respectively. On the regression task, the Feature Explainer significantly outperforms all baselines, obtaining the highest mean AUC for both Fidelity and Deviation metrics. In particular, the Feature Explainer shows superior performance as it is capable of extracting important features which leads to even sparser explanations.

TGNN‑Explainer generally performs worst across all real-world benchmarks. The driving factor is the hard-coded limit of 20 randomly chosen events per instance that are used for the explanation. All other events remain in the computational subgraph leading to very dense explanations. However, it shows a comparable performance on smaller datasets and good generalizability to regression tasks. TempME produces competitive scores, particularly on datasets it was originally designed for (Wikipedia, Reddit), but weaker performance on previously unreported datasets or regression tasks (Generated, MOOC) reflects reduced generalizability. Further, the public code, especially the motif extraction, is incomplete (cf.~\cite{qu2025cody}). To overcome this problem, we had to reconstruct the preprocessing based on the out-commented parts in the initial repository.

\paragraph{Insights from sparsity curves.}
In Fig.~\ref{fig:evalWikipedia} and Fig.~\ref{fig:evalGen}, we exemplary report the Sparsity-vs-Fidelity curves for the Wikipedia dataset and the artificial dataset on a TGAT model. For the generated dataset, the proposed Shapley‑ and Owen‑based explainers achieve steep metric increases at very low sparsities, indicating they quickly identify the most relevant events or features. For the Wikipedia dataset, TempME’s curves remain below the ones of the Shapley explainer which aligns with a higher AUC value of the Shapley explainers. By contrast, TGNN‑Explainer curves remain flat up to sparsities $\approx0.8$, because its minimal explanations exceed that threshold. However, for the removing technique with average events, the TGNN explainer sometimes outperforms the other explainers although its curve stays flat across most sparsities. This is likely due to the presence of certain structures that persist in the computational subgraph if events are not removed but replaced. Further, one can observe a sharp drop in the performance of all explainers already reported in the TempME paper. Using the insights obtained by our Feature explainer (see Sec.~\ref{sec:qualitative-eval}), we can conclude that this drop occurs when nodes have no neighbors in the second hop tempting the model to use information from non-existing neighbors. With increasing sparsity thresholds fewer nodes have no neighbors in the second hop recovering the metric from that drop. Our thesis is also supported by the fact that the drop does not occur when events are not removed but replaced by average events. And the drop also does not occur on one hop models.

\paragraph{Runtime analysis.}
Runtime comparisons on the Wikipedia dataset highlight trade‑offs between accuracy and computational cost. During inference, TempME achieves the shortest per‑explanation generation time ($\approx1.1$ sec.), but requires a pretrained auxiliary explainer network and extensive preprocessing (taking multiple hours), thus incurring high initialization overhead. The Event Explainer reaches comparable inference times ($\approx1.8$ sec.) without training, whereas the Feature Explainer is slower ($\approx50$ sec. per event) due to additional sampling needed for feature-level Owen values. TGNN‑Explainer exhibits the longest inference time ($\approx110$ sec.) owing to repeated MCTS expansions. 

\subsection{Qualitative Evaluation}
\label{sec:qualitative-eval}

\begin{figure}
    \centering
    \includegraphics[width=0.65\linewidth]{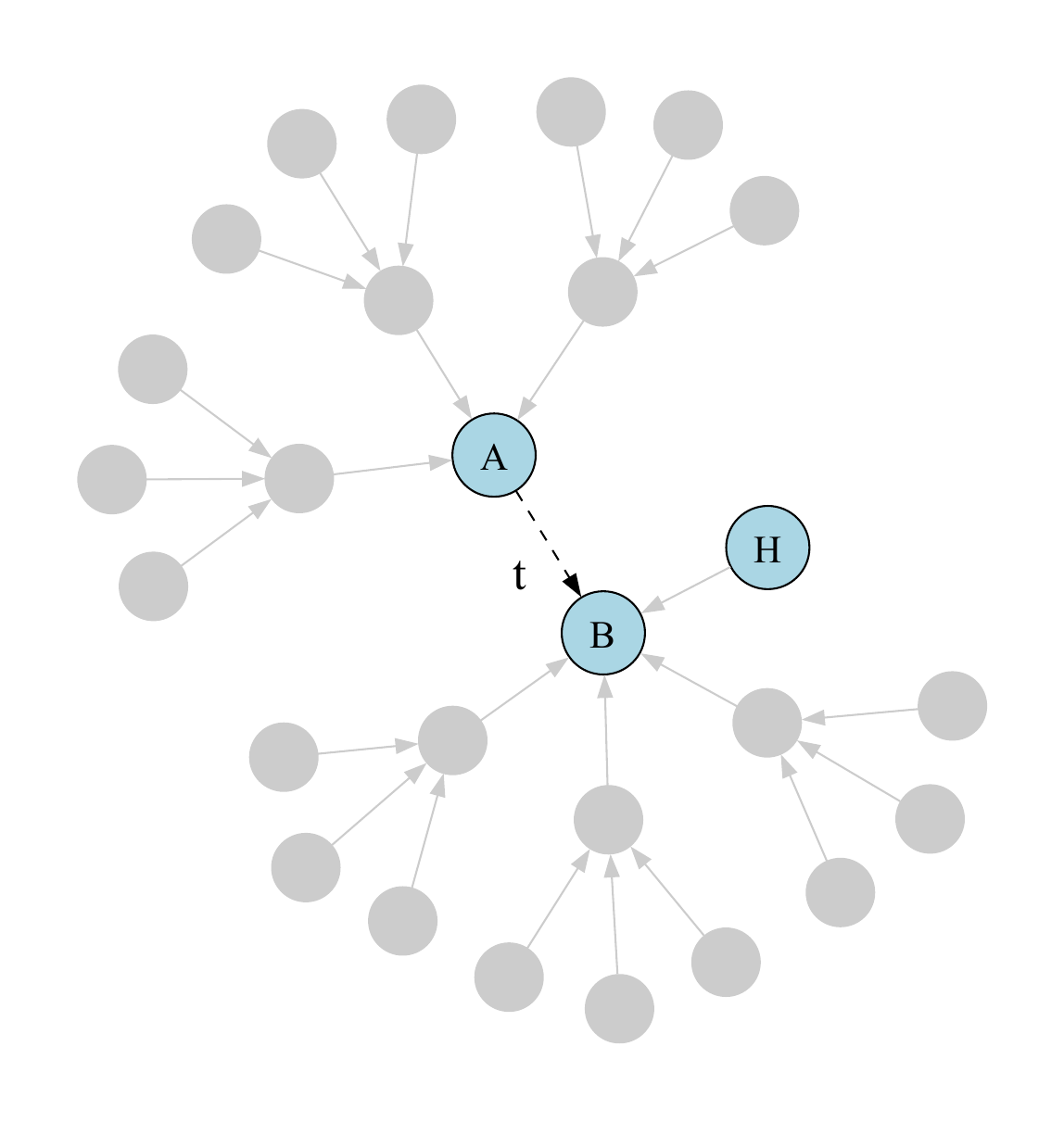}
    \caption{Example of a graph structure within the artificial dataset. The goal is to predict the actual timestamp $t$ using the subgraphs around $A$ and $B$. Each node can have up to three connections occurring at $t-1$ and these can further have up to three connections occurring at $t-2$. Node $B$ always has an event at $t-1$ with node $H$, but $H$ has no further events.}
    \label{fig:exp-art}
\end{figure}

We use a synthetic dataset and a two-layer TGAT model to quantitatively evaluate the proposed \textit{Feature Explainer}. Further, we aim to show an unexpected behavior of TGAT on the artificial dataset. The task is to predict the timestamp~$t$ from the computational subgraphs of nodes~$A$ and~$B$ (see Fig.~\ref{fig:exp-art}). As the TGAT architecture operates on time deltas, which are fixed at~$1$ in this dataset, the model should not be able to predict the absolute timestamp~$t$. Nevertheless, it achieves high predictive accuracy on both training and test sets. Because identical graph structures occur at different timestamps, simple memorization cannot explain this result.

We applied our explainers to generate attributions for sample predictions. The event between nodes~$H$ and~$B$ consistently receives a high Shapley value, despite the constant time delta and event feature. Using our Feature Explainer, we further found that the node features of~$H$ exhibit high Owen values, accounting for most of the event’s attribution. As all node features are initialized to zero, this indicates unintended information propagation toward~$H$.

A closer inspection of the TGAT implementation revealed that this effect originates from the attention mechanism. ``Non-existent'' events are created with node features, event features, and timestamps all set to~$0$. Normally, such events receive very low attention scores, but if all scores are low (i.e., when no actual neighbor exists around the node), the softmax normalizes them uniformly. This results in a time delta of $-(t-1)$, allowing the model to infer the absolute timestamp. Thus, the Feature Explainer successfully identifies the source of this spurious signal, confirming that the generated explanations are both valid and informative.

\section{Discussion}
The Feature Explainer provides fine-grained explanations by decomposing event‑level Shapley values into feature‑level contributions via Owen values. This hierarchical explanation is especially useful when event features have semantic meaning and are human interpretable. However, its longer runtime limits it to analyses on selected events rather than a broad usage. Both explainers combined enable an efficient, and extensive understanding of TGNN behavior.

Compared to TempME, our methods remove the necessity of training a dedicated explainer and the dependence on attention-based TGNNs. TempME can achieve high fidelity on certain benchmarks but requires extensive hyper‑parameter tuning and motif extraction, making it less adaptable and reproducible. The disparities observed between our results and the original reported curves suggest that incomplete motif preprocessing and missing documentation complicate its reproducibility. Although TGNN‑Explainer is a valuable baseline, its performance is limited by its fixed candidate events and its usage of a trained navigator.

Our Shapley-based approach assumes that events within a computational subgraph can be modified independently, although this assumption does not strictly hold. Hence, we explored two ways of ``removing events'': (1) actual removal, which can lead to disconnected graphs and impairs message passing, and (2) replacement with average events, which affects their feature values but leaves the graph structure intact. Our evaluation shows that both variants yield high-quality explanations outperforming the baselines.

\section{Conclusion}

In conclusion, this work introduces two novel explanation methods for Temporal Graph Neural Networks (TGNNs): The Event Explainer based on Shapley values and the Feature Explainer based on Owen values. Both methods are model-agnostic and provide fine-grained explanations at the event and feature level. Experiments across benchmark datasets demonstrate that the new explainers match or surpass existing approaches in several metrics without requiring model‑specific adaptations or additional training. Moreover, the Feature Explainer revealed unexpected leakage of absolute time information within attention‑based TGNNs, underscoring its diagnostic potential.

\section*{Acknowledgements}

The authors gratefully acknowledge the funding of this project by computing time provided by the Paderborn Center for Parallel Computing (PC2). Portions of the paper were refined using an OpenAI language model. The model was used for editing and restructuring text; no scientific conclusions were generated without independent verification by the authors.

{
\raggedright
\bibliographystyle{named}
\bibliography{bibliography}
}
\end{document}